\documentclass[sigconf]{acmart}
\AtBeginDocument{%
  }

\copyrightyear{2025}
\acmYear{2025}
\setcopyright{cc}
\setcctype{by}
\acmConference[SIGIR '25]{Proceedings of the 48th International ACM SIGIR Conference on Research and
Development in Information Retrieval}{July 13--18, 2025}{Padua, Italy}
\acmBooktitle{Proceedings of the 48th International ACM SIGIR Conference on Research and Development in
Information Retrieval (SIGIR '25), July 13--18, 2025, Padua, Italy}\acmDOI{10.1145/3726302.3730344}
\acmISBN{979-8-4007-1592-1/2025/07}
\acmDOI{10.1145/3726302.3730344}

\usepackage{graphicx}

\usepackage{xcolor}         
\usepackage{tikz}
\usetikzlibrary{trees}
\usepackage{soul}

\usepackage{multirow}
\usepackage{multicol}
\usepackage{adjustbox}

\usepackage{subcaption}
\usepackage[most]{tcolorbox}
\usepackage{comment}

\usepackage{booktabs}
\usepackage{hyperref}
\usepackage{wrapfig}
\usepackage{svg}
\PassOptionsToPackage{table}{xcolor}
\definecolor{Emerald}{rgb}{0.0, 0.67, 0.87}
\definecolor{Dandelion}{rgb}{1.0, 0.75, 0.0}
\sethlcolor{yellow!50}

\begin{document}

\title{Wiki-TabNER: Integrating Named Entity Recognition into Wikipedia Tables}


\author{Aneta Koleva}
\affiliation{%
  \institution{Ludwig-Maximilians-Universit\"{a}t}
  \city{Munich}
  \country{Germany} \\
  \institution{Siemens AG}
  \city{Munich}
  \country{Germany}} 
\email{aneta.koleva@siemens.com}

\author{Martin Ringsquandl}
\affiliation{%
  \institution{Siemens AG}
  \city{Munich}
  \country{Germany}} 
\email{martin.ringsquandl@siemens.com}

\author{Ahmed Hatem}
\affiliation{%
 \institution{Technical University of Munich}
 \city{Munich}
  \country{Germany}} 
\email{ahmed.hatem.m.g@gmail.com}

\author{Thomas Runkler}
\affiliation{%
\institution{Technical University of Munich}
 \city{Munich}
  \country{Germany} \\
  \institution{Siemens AG}
  \city{Munich}
  \country{Germany}}    
\email{thomas.runkler@siemens.com}

\author{Volker Tresp}
\affiliation{%
  \institution{Ludwig-Maximilians-Universit\"{a}t}
  \city{Munich}
  \country{Germany} \\
  \institution{Siemens AG}
  \city{Munich}
  \country{Germany}}  
\email{volker.tresp@lmu.de}

\renewcommand{\shortauthors}{Aneta Koleva, Martin Ringsquandl, Ahmed Hatem, Thomas Runkler, and Volker Tresp}

\begin{abstract}
  Interest in solving table interpretation tasks has grown over the years, yet it still relies on existing datasets that may be overly simplified. This is potentially reducing the effectiveness of the dataset for thorough evaluation and failing to accurately represent tables as they appear in the real-world. To enrich the existing benchmark datasets, we extract and annotate a new, more challenging dataset. The proposed Wiki-TabNER dataset features complex tables containing several entities per cell, with named entities labeled using DBpedia classes. This dataset is specifically designed to address named entity recognition (NER) task within tables, but it can also be used as a more challenging dataset for evaluating the entity linking task. In this paper we describe the distinguishing features of the Wiki-TabNER dataset and the labeling process. In addition, we propose a prompting framework for evaluating the new large language models on the within tables NER task.
Finally, we perform qualitative analysis to gain insights into the challenges encountered by the models and to understand the limitations of the proposed~dataset.
\end{abstract}



\maketitle

\section{Introduction}

Representing complex data in tables enhances readability and facilitates improved data comprehension. The abundance of web tables \cite{WebTables} and the recent advances in natural language processing (NLP) pioneered by the transformer model \cite{attentio_is_all_you_need} have been the inspiration behind the numerous newly proposed tabular language models \cite{mate, tapas, tabbie, tuta, tabert, table2vec}.
These models are evaluated for solving table interpretation (TI) tasks.
These tasks aim at discovering the semantics of the data captured in the tables and include: entity linking (EL), where the objective is to link entity mentions from table cells to reference entities; column type annotation (CTA), where columns are annotated with semantic types; and relation extraction (RE), where the semantic relations between columns are identified. 

\begin{figure*}[t!]
\centering
\begin{minipage}[t]{0.33\textwidth}
    \centering
    \includegraphics[width=\linewidth]{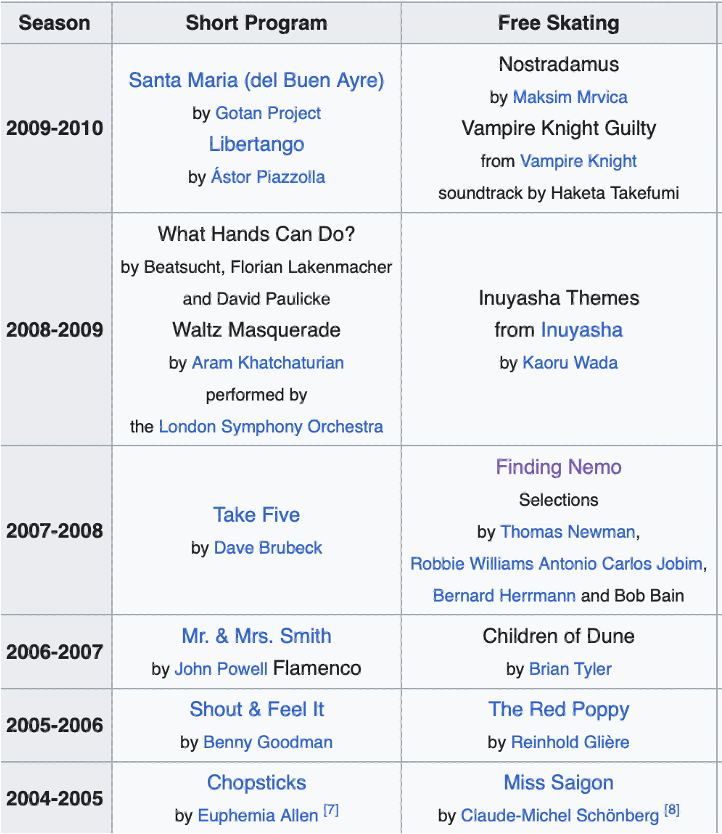}
\end{minipage}%
\hspace{0.001\textwidth} 
\begin{minipage}[t]{0.42\textwidth}
    \centering
    \vspace{-6.77cm} 
    \begin{adjustbox}{width=\linewidth}
    \begin{tabular}{|l|c|c|}
    \toprule
    \textbf{Season} & \textbf{Short Program} & \textbf{Free Skating} \\
    \hline
    2009-2010 & \textcolor{red}{Santa Maria (del Buen Ayre)} & Nostradamus by \textcolor{Dandelion}{Maksim Mrvica} \\
              &  by \textcolor{Dandelion}{Gotan Project}  & \textcolor{red}{Vampire Knight Guilty} from \\
              & \textcolor{red}{Libertango} by Åstor Piazzolla & Vampire Knight \\ 
               & & soundtrack by Haketa Takefumi \\
    \hline
    2008-2009 & What Hands Can Do? by Beatsucht, & Inuyasha Themes \\
              & Florian Lakenmacher and & from \textcolor{red}{Inuyasha} \\
              & Paulicke & by \textcolor{Emerald}{Kaoru Wada} \\
              & Waltz Masquerade by & \\
              & \textcolor{Emerald}{Aram Khatchaturian} performed by & \\
              & the \textcolor{Dandelion}{London Symphony Orchestra} & \\    
    \hline
    2007-2008 & \textcolor{red}{Take Five} by \textcolor{Emerald}{Dave Brubeck} & \textcolor{red}{Finding Nemo} Selections \\
                & & by \textcolor{Emerald}{Thomas Newman}, \textcolor{Emerald}{Robbie Williams},\\
                & & Antonio Carlos Jobim, \textcolor{Emerald}{Bernard} \\
                & & \textcolor{Emerald}{Herrmann}, and Bob Bain \\
    \hline
    2006-2007 & \textcolor{red}{Mr. \& Mrs. Smith} by & Children of Dune by\\
                & \textcolor{Emerald}{John Powell} Flamenco & \textcolor{Emerald}{Brian Tyler} \\
    \hline
    2005-2006 & \textcolor{red}{Shout \& Feel It} by & \textcolor{red}{The Red Poppy} by\\
               &  \textcolor{Emerald}{Benny Goodman} &  \textcolor{Emerald}{Reinhold Gliere} \\
    \hline
    2004-2005 & Chopsticks by  & \textcolor{red}{Miss Saigon} by \\
               & \textcolor{Emerald}{Euphemia Allen} &  \textcolor{Emerald}{Claude-Michel Schönberg} \\
    \bottomrule

  \end{tabular}

  \end{adjustbox}
\end{minipage}%
\hspace{0.001\textwidth} 
\begin{minipage}[t]{0.23\textwidth}
    \centering        
    \vspace{-6.77cm} 
    \begin{adjustbox}{width=\linewidth}
    \begin{tabular}{|c|c|}
    \hline
    \textbf{Short Program} & \textbf{Free Skating} \\
       \hline
              & Maksim Mrvica   \\         
              \hline
         Aram Khatchaturian & Inuyasha  \\  
             \hline
         Take Five & Finding Nemo \\
             \hline
         Mr. \& Mrs. Smith & Brian Tyler\\
             \hline
               & The Red Poppy \\
             \hline
               Chopsticks & Miss Saigon\\
            \hline
    \end{tabular}
    \end{adjustbox}
    \label{tab:my_label}
\end{minipage}
\caption[]{Complex table retrieved from Wikipedia (left). Representation of the original table in the Wiki-TabNER dataset (center) and in the TURL dataset (right). The Wiki-TabNER table contains named entities of type: \textcolor{red}{Work}, \textcolor{Emerald}{Person}, and \textcolor{Dandelion}{Organization}.}
  \label{fig:complex_table}
\end{figure*}
There are several datasets commonly used for the evaluation of the TI tasks, T2D \cite{t2d_2015,t2dv2} and Limaye \cite{limaye} which consist of small number of manually annotated tables, Sato \cite{sato} and GitTables \cite{gitTables} which contain tables annotated for the CTA task and the TURL \cite{turl} dataset with a large collection of tables annotated for all of the TI tasks. Nevertheless, all of these datasets either constrain the text within each cell to contain at most one entity mention \cite{limaye,t2dv2,turl}, or assume that all entities within a cell are of the same semantic type as their column \cite{sato, gitTables}. 

The original table in Figure \ref{fig:complex_table} illustrates that tables in reality are more complex. Often, tables in the real-world contain several entities per cell and additional text surrounding the entities \cite{koleva2022named}. In order to extract entities from tables and link them to a knowledge base for the EL task, it is essential to first identify the entities within the cells. Therefore, if we discard the single-entity-per-cell assumption and consider cell text without removing any content, including other entities and non-entity tokens (e.g., ``by'', ``and'', etc.), we first need to perform NER within cell to enable EL. This highlights the need to integrate the task of NER within tables.

Thus, to complement the landscape of existing datasets, we introduce a novel dataset, Wiki-TabNER, which reflects the Wikipedia tables with their real structure. Moreover, we have labeled the entities within the cells with $7$ DBpedia entity types \cite{dbpedia} with the intention to utilize this dataset for evaluating the NER task within tables. In Figure \ref{fig:complex_table}, we show an example of how the original table from Wikipedia looks like, how the same table is presented in the novel Wiki-TabNER dataset and how this table is in the existing TURL dataset \cite{turl}. 
Observing this example, we see the discrepancy between the original table and the simplified table that is used for evaluation of TI tasks. In the Wiki-TabNER dataset we annotate named entities using span-based labels \cite{spanbert}. This allows for an evaluation for the NER task within tables with transformer based models. 

Even though NER is a long-standing problem in the NLP community, this problem so far has not been observed within tables. NER in a single table cell is fundamentally similar to natural language NER; however this assumption only holds if the task is considered independently of other cells in the same table. The challenge lies in correctly interpreting the relationships between a table's schema, it's rows and columns and the content within the cells. Table NER can be defined as follows: Identify all entity mentions in a cell and classify each entity into a semantic type. In this paper we extend the idea of table NER to any relational table, not only to industrial tables as in \cite{koleva2022named}. We first show the difference to the existing datasets, then we present the construction of the new benchmark dataset Wiki-TabNER. Following the trend of increasingly using LLMs for solving various tabular problems \cite{tablegptHeng, TableGPTZha, tablellama}, we also propose an evaluation framework for in-context learning of LLMs on the Wiki-TabNER dataset. We explore the capabilities of these models for NER in tables by conducting experiments in zero, one and few shot settings. To the best of our knowledge, this is the first work to propose a benchmark dataset with multi-entity cells which is closer to real-world use cases. We release the proposed dataset, alongside guidance for usage and the evaluation framework for LLMs online\footnote{\url{https://github.com/table-interpretation/wiki_table_NER}}.

\begin{table*}
\centering
    \caption{Characteristics of the existing datasets and Wiki-TabNER. The * indicates that these are subsets of larger datasets. The higher average number of tokens per cell in Wiki-TabNER indicates that these tables contain significantly denser information compared to the other datasets.}
    \label{dataset_stats}
   \centering
    \begin{tabular}{c c c c c c c}
    \toprule
     \textbf{Dataset} & \textbf{\# tables} & \textbf{$\mu$ rows}  & \textbf{$\mu$ columns}  &  \textbf{$\mu$ tokens} & \textbf{$\mu$, split}\\ 
     \midrule
      T2D \cite{t2d_2015}  & 233 & 112.6\footnotesize $\pm$ 115.5  & 4.9\footnotesize$\pm$ 1.8 & 1.9\footnotesize$\pm$ 1.8 & 1\footnotesize$\pm$ 0  \\
      Limaye \cite{limaye} & 428 & 34.5\footnotesize$\pm$ 41 & 3.8\footnotesize$\pm$ 1.2 & 1.7\footnotesize$\pm$ 1.4 & 1\footnotesize$\pm$ 0.2 \\
      WikipediaGS \cite{matchingwebtables} & 15000* & 13.8\footnotesize$\pm$ 35.5 & 4.6\footnotesize$\pm$ 3.4 & 2.5\footnotesize$\pm$ 5.7 & 1.1\footnotesize$\pm$ 0.6\\      
      TURL(EL) \cite{turl} & 10927*  & 17.5\footnotesize$\pm$ 13.6  & 2.8\footnotesize$\pm$ 1 & 1.5\footnotesize$\pm$ 0.9  &  1\footnotesize$\pm$ 0\\
      \textbf{Wiki-TabNER} & 61235 & 12.7\footnotesize$\pm$ 19.9  & 5.2\footnotesize$\pm$ 2.4 & \textbf{4.6}\footnotesize $\pm$ 16.1 & 1.2\footnotesize$\pm$1.4\\
      \bottomrule
    \end{tabular}    
\end{table*}
\section{Characteristics of Table Benchmark Datasets}\label{related work}

Several benchmarks have been proposed for the evaluation of the TI tasks. In Table \ref{dataset_stats} we show an overview of their characteristics. For each dataset, we show the total number of tables and the mean and standard deviation of the number of rows and columns. Additionally, using the Spacy tokenizer\footnote{\url{https://spacy.io/api/tokenizer}}, we tokenize the cell text and calculate the mean and standard deviation of the number of tokens per cell in the tables. The last column shows the mean number of entities per cell, based on splitting the cell text using the comma delimiter.

\paragraph{T2D} Originally published in 2015, the T2D dataset \cite{t2d_2015} consists of manually annotated row-to-entity correspondences, between $233$ Web tables and instances from the DBpedia knowledge base \cite{dbpedia}. Column annotations and table-to-class annotations were also provided. These tables were extracted from the English-language subset of the Web Data Commons Web Tables Corpus \cite{WDC}. The dataset was extended to T2Dv2 \cite{t2dv2} with $779$ tables, out of which $237$ tables have entities linked to DBpedia instances. In Table \ref{dataset_stats} we observe that even though this dataset has on average the largest tables, the cell text has been cleaned so that it contains only one~entity~per~cell.

\paragraph{Limaye} Another benchmark was proposed by Limaye et al. \cite{limaye} which contains 437\footnote{This is the number of tables that is reported in \cite{limaye}. In our calculation we used the dataset provided here \url{http://www.cs.toronto.edu/~oktie/webtables/}, which has 9 tables less.} cell-level and column-level manually annotated tables using Wikipedia, DBpedia and YAGO \cite{yago}. This dataset is used for the evaluation of the EL, CTA and RE tasks. The tables are extracted with queries and manually annotated, resulting in a dataset with clean, simple tables.

\paragraph{WikipediaGS} As a more challenging dataset, Efthymiou et al. \cite{matchingwebtables} created the Wikipedia gold standard (WikipediaGS) benchmark. This dataset consists of 485,096 tables from Wikipedia and is intended for the task of matching rows to DBpedia entities. We randomly selected a subset of 15000 tables to calculate the statistics in Table \ref{dataset_stats}, which correspond to the statistics of the overall dataset as presented in \cite{matchingwebtables}. This dataset has not been pre-processed and it contains tables with unstructured, free-form text within the cells. This is also visible in Table \ref{dataset_stats} in the mean number of tokens per cell.

\paragraph{TURL (EL)} The TURL dataset \cite{turl} is an established benchmark dataset for all of the TI tasks. It is a large dataset, extracted from the WikiTable corpus \cite{tabel}, with more than 500k tables which can be used for pre-training and fine tuning of models aiming to solve various tabular tasks. For the calculation of the statistics we used the subset of dev and test tables annotated for the EL task. The tables in this dataset are pre-processed so that the cells contain only one entity, the rest of the text is discarded. In Figure \ref{fig:complex_table}, we can see the discrepancy between the original Wikipedia page and the cleaned table in the TURL dataset. The statistics in Table \ref{dataset_stats} also show that, the tables in this dataset have the least number of columns and that the average number of tokens per cell is only 1.5.

\paragraph{Wiki-TabNER dataset} In Table \ref{dataset_stats} we also show the statistics of the new dataset, Wiki-TabNER. Compared to the other datasets, the distinguishing feature of this dataset is the higher average number of tokens per cell. Unlike existing datasets, we do not preprocess the cell text, and we extract tables containing multiple entities per cell. Consequently, the tables contain a lot of \textit{noise} (unstructured text) surrounding the entities. Furthermore, the last column in Table \ref{dataset_stats} reveals that all datasets have an average of 1.1 entities per cell when using a comma as a delimiter to separate the entities. However, in the Wiki-TabNER tables, cells can contain multiple entities alongside non-entity tokens (such as: `by', `and', `from', etc.) as illustrated in Figure \ref{fig:complex_table}. We calculated that the average number of labeled entities per cell in the Wiki-TabNER dataset is actually 2.0. Therefore, a simple heuristic of splitting entities on a comma delimiter is not sufficient for identifying the entities within the cells. This underscores the necessity for more sophisticated methods to accurately identify the entities within the cells. 

Additionally, to quantify the significance of the difference in the average number of tokens per cell between the Wiki-TabNER and WikipediaGS \cite{matchingwebtables} datasets, we conducted the Welch's t-test \cite{welch}. The result (T-statistic=223.89, p-value=0) confirms a statistically significant difference between the mean number of tokens in these two datasets, thus reinforcing the assertion of denser information content in the Wiki-TabNER dataset. 

\section{Dataset Construction} \label{wikitabner_dataset}
Motivated by the absence of a dataset containing tables which resemble the real-world tables among the existing benchmark datasets and by the potential of the WikiTables corpus \cite{tabel}, we propose a new benchmark dataset consisting of tables with denser information. The presented dataset can be used for evaluation of within-table NER and EL.

\subsection{Extracting tables from the WikiTables corpus}
Following the creation of the TURL dataset \cite{turl}, we extract a high quality subset of relational tables from the WikiTable corpus \cite{tabel}, by identifying tables that have a subject column. The subject column must be located in the first two columns of the table and contains unique entities which are treated as subject entities. 
Moreover, with the intention to create a dataset with complex tables we only retain tables that have: at least 3 rows, 3 columns and at least one column where 90\% of the cells have an average of at least 2 entities with surface links. We further filter out the tables with undesirable captions, such as ``External Links'', ``References'' and ``Sources''. The resulting dataset is consisting of $62395$ tables. Throughout the labeling process, we remove $1160$ tables for which none of the linked entities could be labeled. Consequently, the final Wiki-TabNER dataset consists of $61235$ NER labeled tables.

\subsection{Labeling Wiki-TabNER}
In tables like those in the Wiki-TabNER dataset, where cells may contain multiple entities along with non-entity tokens, identifying which entities can be linked is a complex task. Hence, solving NER as a first step can help in addressing the EL task. In this direction, we label the linked entities in the Wiki-TabNER dataset with entity types.

\paragraph{1. Extracting linked entities from the tables} 

The tables, as provided in the WikiTables corpus \cite{tabel}, already contain additional information for the entities, including surface links to existing Wikipedia pages. As a first step, we extract all entities from the dataset that have a surface link to a Wikipedia entity. For each linked entity, we collect the following details: the surface link to its corresponding Wikipedia page, the target text of the entity mention in the cell, and the start and end positions of the entity mention within the cell text.

\paragraph{2. Annotating named entities with their semantic types and linking them to Wikidata IDs} To annotate the extracted entities with named entity types, we use the mapping from the entities to their DBpedia \cite{dbpedia} and to Yago \cite{yago} semantic types. This mapping is available in the instance files provided by DBpedia\footnote{\url{https://databus.dbpedia.org}}, where every entity is annotated with a semantic class. 
The DBpedia classes\footnote{\url{https://dief.tools.dbpedia.org/server/ontology/classes/}} are structured in a hierarchical tree with different levels of subclasses. Our goal is to extract the most general classes for each entity, i.e., we annotate the entities with the classes that are direct subclass of \textit{owl:Thing}. We show the classes we used for annotation in Figure \ref{fig:dbpedia-tree}. Assigning entities to their most general class provides them with a broader semantic context and makes it easier to classify them.

To find the classes for the extracted entities, we use the entity target text, which is a string representing the entity. For example, the target text for the entity ``Finding Nemo'' is simply ``Finding\_Nemo'' and its DBpedia class is \textit{Movie}. After mapping the entities to their classes, we then propagate each class to its corresponding most general superclass. For example, we annotate the entity ``Finding Nemo'' as type \textit{Work}, because \textit{Work} is the highest superclass encompassing the \textit{Movie} class.

We observed that the majority of the entities were classified into one of the following classes \textit{{Activity, Agent, Architectural Structure, Event, Place, Species, Work}}. The rest of the classes had significantly fewer entities assigned to them, so we decided to focus on these 7 classes. A closer inspection of the class \textit{Agent} revealed that almost all instances of this class, were actually instances of the class \textit{Organization}, so we substitute the class \textit{Agent} with the more specific one \textit{Organization}. Similarly, all instances labeled as \textit{Species}, were instances of \textit{Person}, so we opted for \textit{Person} instead. 

\begin{figure}[h]
    \centering
    \includegraphics[width=\linewidth]{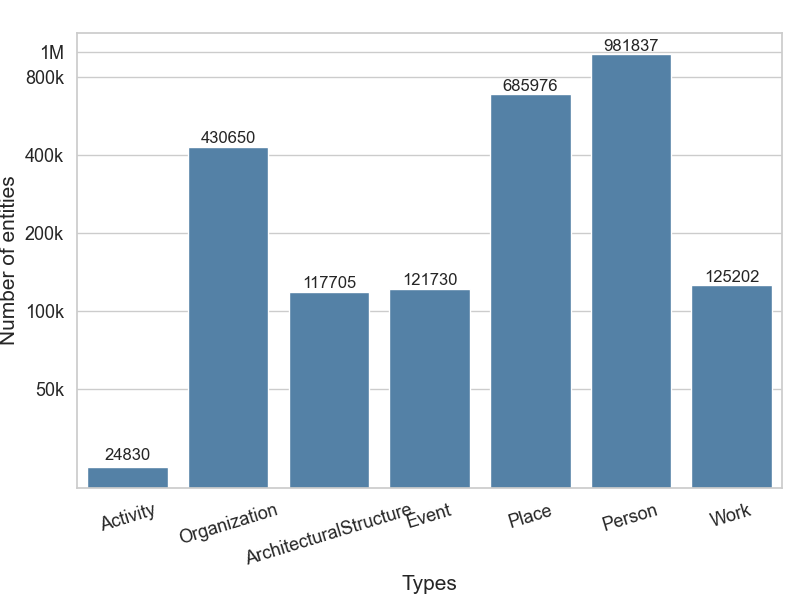}
      \caption[]{Number of entities per type in Wiki-TabNER.}
      \label{fig:num_ent_per_type} 
\end{figure}

Finally, using the WikiMapper\footnote{\url{https://github.com/jcklie/wikimapper}} we map the named entities to their corresponding Wikidata IDs (known as \textit{Q-identifiers}). With this, we create a mapping for the entities to their semantic types and to their Wikidata IDs. Consequently, the annotated Wiki-TabNER dataset can be used not only for evaluating NER within tables but also for the evaluation of the EL task. In Figure \ref{fig:num_ent_per_type}, we show the final distribution of named entities in the dataset. Despite the \textit{Person} class having the highest number of instances, it's noteworthy that in the majority of tables, there are named entities of three or four unique labels. This observation suggests a level of diversity in the types of entities represented across the tables in the dataset. 

\paragraph{3. Generating the span-based labels of the entities and adding them to the table structure}
To facilitate the evaluation of NER models on the Wiki-TabNER dataset, we provide the span-based labels of the entities. 
The span label for an entity includes the cell position, the start and end of the span and the unique integer identifier of the label. We use the information about the start and end positions of the linked text within the cells as the start and end of the entity span. For example the entity ``Finding Nemo'' from the example table in Figure \ref{fig:complex_table} has a span label $[2,2,0,12,7]$ where $2,2$ indicates this entity is in the 2nd row, 2nd column in the table, the start of the span is at position $0$ in the cell and it ends at $12$ and it has label $7$ which corresponds to the entity type \textit{Work}. 
We add the span-based labels for the linked entities as additional information to the table structure.

The resulting dataset contains $408 781$ distinct entities, linked to their Wikidata ID and annotated with a DBpedia or Yago semantic class.
Hence, the labeling of the dataset results in a richly annotated and reusable dataset where each table contains linked entities with different labels. 

However, there are also entities which are linked to a Wikipedia page, but do lack a corresponding DBpedia entry, resulting in a missing entity type. An example is the entity ``Astor Piazzolla'' from the example table in Figure \ref{fig:complex_table}. Additionally, certain entities are not linked at all, so we cannot assign them any entity type. These unlinked entities, such as ``Nostradamus'' and ``Haketa Takefumi'', are disregarded when addressing the NER or EL task. We discuss more about the limitations of the dataset in section \ref{issues}.

\begin{figure}
    \centering   
    
\resizebox{\linewidth}{!}{%
\begin{tikzpicture}[
  sibling distance=5.5em,
  level distance=2em,
  edge from parent/.style={draw, -latex}]
  
  \node {owl:Thing}
    child {node {\hl{Activity}}}
    child {node {Agent}
      child {node {\hl{Organization}}}
    }
    child {node {\hl{ArchStruct}}}
    child {node {\hl{Event}}}
    child {node {\hl{Place}}}
    child {node {Species}
      child {node {Eukaryote}
      child {node {Animal}
      child {node {\hl{Person}}}
    }}}
    child {node {\hl{Work}}};
\end{tikzpicture}%
}
 \caption{Part of the first level of the DBpedia class tree. We highlight the classes with which we annotate the entities in the Wiki-TabNER dataset.}
    \label{fig:dbpedia-tree}
\end{figure}

\begin{figure*}
\small
  \centering
  \begin{tcolorbox}[width=0.98\textwidth,colback=white,colframe=black]
    \color{gray!80}
    \textit{Instruction:} \\
    \color{red!80}
    You are an NER expert. Extract entities from the input table using the following types: Activity, \textbf{Organisation}, ArchitecturalStructure, Event, \textbf{Place}, \textbf{Person}, \textbf{Work}. 
   If the type of the entity is not one of the types above, please use type: MISC. The output is a list with dictionary for every entity in the following format: 
   \[\{``entity": Entity, ``type": Type, ``cell\_index": [x,y]\}\] Cell index should be one list $[x,y]$ where $x$ is the row number and $y$ is the column number.
   The table header has index -1, the table content with entities start from index $[0,0]$. 
   
\begin{minipage}{0.48\textwidth}
\color{gray!80}
    \textit{Example:} \\
\color{blue!80}
    \textbf{Table:} \\
      | Rider | Team | Time \\
    1 | Giorgia Bronzini ( ITA ) | Wiggle-Honda | \\
    2 | Emma Johansson ( SWE ) | Orica-AIS | s.t. \\
\color{blue!80}
\textbf{Output:} \\
    $[\{\text{``entity'': ``Giorgia Bronzini'', ``type'': ``Person'', ``cell\_index'': [0, 1]}\}, \\
    \{\text{``entity'': ``ITA'', ``type'': ``Place'', ``cell\_index'': [0, 1]}\}, \\
    \{\text{``entity'': ``Emma\space Johansson'', ``type'': ``Person'', ``cell\_index'': [1, 1]}\}, \\
    \{\text{``entity'': ``SWE'', ``type'': ``Place'', ``cell\_index'': [1, 1]}\} ]$\\   
\end{minipage}
\hfill
\begin{minipage}{0.48\textwidth}    
\vspace{-1cm} 
 \color{gray!80}
    \textit{Input Table:}\\    
    \color{violet!80}
    \textbf{Table:} \\    
      | Rider | Team | Time \\
    1 | Giorgia Bronzini ( ITA ) | Wiggle-Honda | 1h 57' 41"\\
    2 | Rosella Ratto ( ITA ) | Hitec Products UCK | s.t.\\
    3 | Ashleigh Moolman ( RSA ) | Lotto Belisol Ladies | s.t.\\
    4 | Karol-ann Canuel ( CAN ) | Vienne Futuroscope | s.t.\\
\end{minipage}
  \end{tcolorbox}
  \caption{Example of a prompt with one-shot example. The instructions part is in red. One example table and its named entities are in blue. The input table for annotation is in violet.}
  \label{fig:prompt}
\end{figure*}

\section{Table NER with LLMs}
We now present how the Wiki-TabNER dataset can be used to evaluate the performance of LLMs on the NER task within tables.

A table defined by $T = (C,H)$ stores information in a 2 dimensional arrangement with $n$ rows and $m$ columns where $C = \{c_{1,1}, c_{1,2}, ..., c_{n,m}\}$ is the set of body cells, and $H = \{h_1, h_2, ..., h_m\}$ is the set of table headers. Each cell $c_{i,j}$ is composed of a list of $t$ tokens: $c_{i,j} = (w_{c_{i,j},1}, w_{c_{i,j},2}, ..., w_{c_{i,j},t})$. The task is to accurately assign an entity type to all of the tokens within a cell. The sequence of entity types is denoted as $Y=\{y_1, y_2, ..., y_n\}$, where each $y_i$ represents a specific entity type. 
Solving the NER task in tables using LLMs consists of several steps. 

\paragraph{1. Input Prompt} \label{sec:input_prompt}

Figure \ref{fig:prompt} shows an example of an input prompt. The instruction part of the prompt describes the task and it instructs the model how the output should be formatted. It is always the same for all of the models and under the different settings. The example part of the prompt provides a \textit{k}-shot example to the model, where $k \in \{0, 1, 3\}$. Each example consists of a table and an output with the annotated entities from this table. 
Finally, the last part of the prompt is the input table from which the model should extract the entities and identify their entity types. 

\paragraph{2. Completion of the prompt}
The generated input prompt is forwarded to an LLM using the completion API.
In the second step, the LLM generates the completion to the input prompt by assigning entity types to the recognized entities in the table and structuring the output. If the model's output does not adhere to the specified format, we save the output to a log file for subsequent analysis.

\paragraph{3. Extracting span-based predictions}
The generated output of the model is assumed to be of the same format as the \textit{Output} from the example part in the prompt. In order to evaluate it, we process the output as follows: first, we serialize the generated output into JSON format, where every annotated entity is a separate JSON entry. Then, for every entity, we extract the entity text, the entity type and its cell position. To find the correct span of the entity in the table, we search in the input table, at the cell position if there is an exact match with the entity text. If yes, we extract the start and end of the entity text within the cell. Finally, we convert the predicted entity type into a numerical value and represent it with a tuple of $5$ elements. In the tuple $(x, y, i, j, k)$, $x$ and $y$ represent the cell index, $i$ and $j$ indicate the span position of the entity in the cell, and $k$ is predicted the entity type.

For example, the parsed output for the first $2$ rows from the example table in Figure \ref{fig:prompt} are : \\ $[(0, 1, 0, 16, 6), (0, 1, 19, 22, 5), (1, 1, 21, 24, 5)]$. 
Indeed, the entity ``Giorgia Bronzini'' is in the cell in the 0th row, column $1$, the span of the text is from $0$ to $16$ and the type of the entity is \textit{Person} which has a corresponding numerical value $6$. In the same cell with index $(0,1)$ is also the entity ``ITA'' which has the span $(19,22)$ and is of type \textit{Place}, represented with the numerical value $5$.

\section{Evaluation}
\begin{figure}
\begin{subfigure}[!t]{0.55\textwidth}
    \hspace{-0.5cm}
    \includegraphics[width=\textwidth]{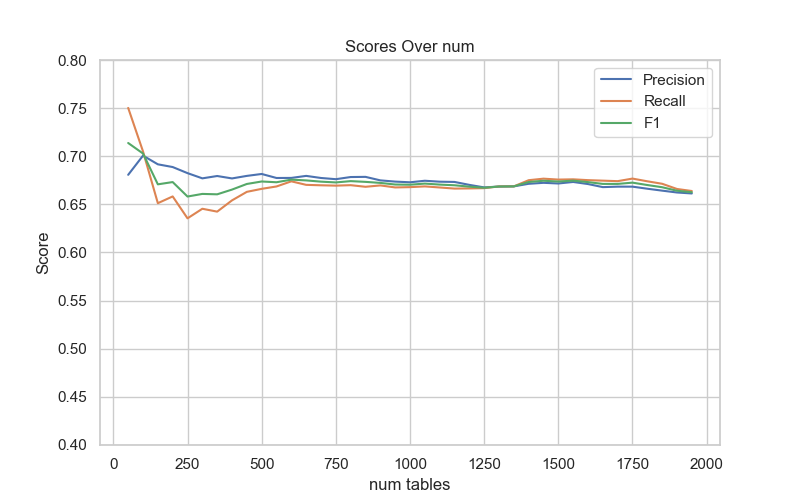}   
    \label{fig:conf}
    \caption{}
\end{subfigure}

\begin{subfigure}[!t]{0.3\textwidth}
\resizebox{!}{0.26\textwidth}{
\small
\hspace{-1.1cm}
    \begin{tabular}{c c c c c }
    \toprule
       & \multicolumn{2}{c}{\textbf{2000 tables}} & \multicolumn{2}{c}{\textbf{600 tables}}  \\
        \midrule
        Model & F1-score & Time &  F1-score & Time \\
        \midrule
        GPT-instruct & 0.50 & 04h50m & 0.51   & 01h10m \\
                
        GPT-3.5-turbo & 0.40 & 04h19m & 0.40 & 01h30m\\
        GPT-4  & 0.41  & 28h34m & 0.41 & 11h30m\\
        \bottomrule
    \end{tabular}
    }
    \caption{}
    
\end{subfigure}

\caption{Results from initial experiments with 2000 evaluation tables. The plot in Figure (a) shows the performance of the GPT-instruct model and we see that there is no change in the metrics after the 600 table. The Table (b) shows the significant amount of duration of these experiments.}
\label{tab:time_execution}
\end{figure}

Our evaluation aims to demonstrate the application of the proposed dataset for within table NER task, using LLMs. The intent is to highlight the challenging nature of the dataset but also to uncover its limitations. 
To achieve this, we concentrate on the span-based predictions extracted by the model. This process includes a comparison between the sequence of tuples that represent annotated entities in the ground truth and the sequence of tuples generated by the LLMs. Our goal is to determine the quantity of exact matches present in these two sequences. 
To quantify the performance of our model, we use the F1-score as evaluation metric. 
\paragraph{\textbf{Dataset}}
For the evaluation we use the Wiki-TabNER dataset, described in Section \ref{wikitabner_dataset}. Although models are currently not trained or fine-tuned on this dataset, the dataset is split into train and test sets for future model training. Additionally, we use the train set to sample k-shot examples when prompting the LLMs. During evaluation, we provide a complete table from the test set as input to the model and we use the ground truth annotations for assessment. Initially, we conducted experiments on a test set of randomly chosen $2000$ tables. However, these experiments took a considerable amount of time to complete, and we noticed that there is no significant change in the metrics after the 600th table. Hence, we decided to evaluate on a smaller test set consisting of $600$ tables. We show the execution times and results from the initial experiments in Figure \ref{tab:time_execution}.

\paragraph{\textbf{Models}}
We conducted the experiments with the Open-AI LLMs\footnote{\url{https://platform.openai.com/docs/models}}: GPT-instruct, GPT-3.5-turbo-16k \cite{Brown2020} and GPT-4 model \cite{gp4}. We also evaluated the open source model Llama2-7b \cite{llama2}. Additionally, we tried evaluating the TableLlama model \cite{tablellama} but this model was struggling with the task and only outputting the instruction part, so we were unable to get any meaningful output for evaluation.

\begin{table}
\caption{F1-score with k-shot examples sampled at random and by similarity.}
\label{tab:rand_sim}
\centering
    \begin{tabular}{c c c c c c c}
    \toprule
       & 
       & \multicolumn{2}{c}{\textbf{Random}} & \multicolumn{2}{c}{\textbf{Similarity}} & \\
        \midrule
        \textbf{Model} & 0-shot & 1-shot & 3-shot  & 1-shot & 3-shot   \\
         \midrule
       GPT-instruct & 0.51 & 0.51 & 0.54 & 0.67 & 0.69   \\
       GPT-3.5-turbo & 0.40 & 0.41 & 0.48 & 0.68 & 0.73  \\
       GPT-4 & 0.41 & 0.43 & 0.46 & 0.68 & \textbf{0.74} \\
       Llama2-7b & 0.02 & 0.04 & 0.08 & 0.23 & 0.29 \\
       \bottomrule
    \end{tabular}
\end{table}

\paragraph{\textbf{Results}}
In Table \ref{tab:rand_sim} we summarize the results and we show the achieved F1-score for all of the models. 
To explore the impact of the few-shot examples, we evaluate two different strategies for sampling the examples: random selection and similarity-based selection. 

For the random selection, we sampled at random three tables from the train set and we extracted their NER annotations. For the one-shot experiment we use one of these tables as an example. To be consistent, we use the same example tables for all of the test tables and for all of the models. 

In contrast to selecting examples at random, we also explore the strategy of selecting examples that are similar to the input table. The rationale behind this approach is that similar examples are likely to provide a more accurate and contextually relevant response, thereby improving the efficiency and effectiveness of the LLM prompting. 
For the similarity-based selection of examples, we first compute the contextual vector for every table in the Wiki-TabNER dataset. 
We compute the BERT \cite{bert} contextual vector for every linearized table where we concatenate the rows into one string. To find similar tables, for every table in the test set, we compute the dot product with every table in the train set. 
During the evaluation, based on the similarity score, we sample either one or three most similar tables as k-shot examples. In Figure \ref{fig:prompt} the example table is chosen based on high similarity to the input table. 

It is evident that all of the models improve with the addition of example to the input prompt. As expected, the similarity-based sampling strategy improves the results by a larger margin than the random-based strategy.   

\begin{figure*}[htbp]
    \centering
    \begin{subfigure}[b]{0.3\textwidth}
        \hspace{-10pt}
        \includegraphics[width=\textwidth]{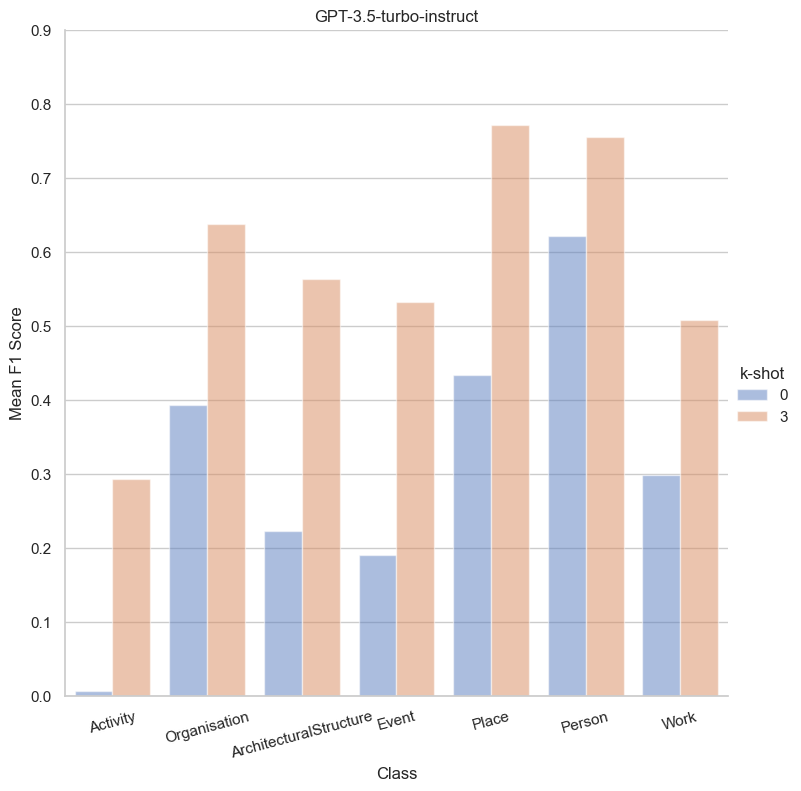}             
    \end{subfigure}
    \hspace{-1pt}
    \begin{subfigure}[b]{0.3\textwidth}
        \centering
        \includegraphics[width=\textwidth]{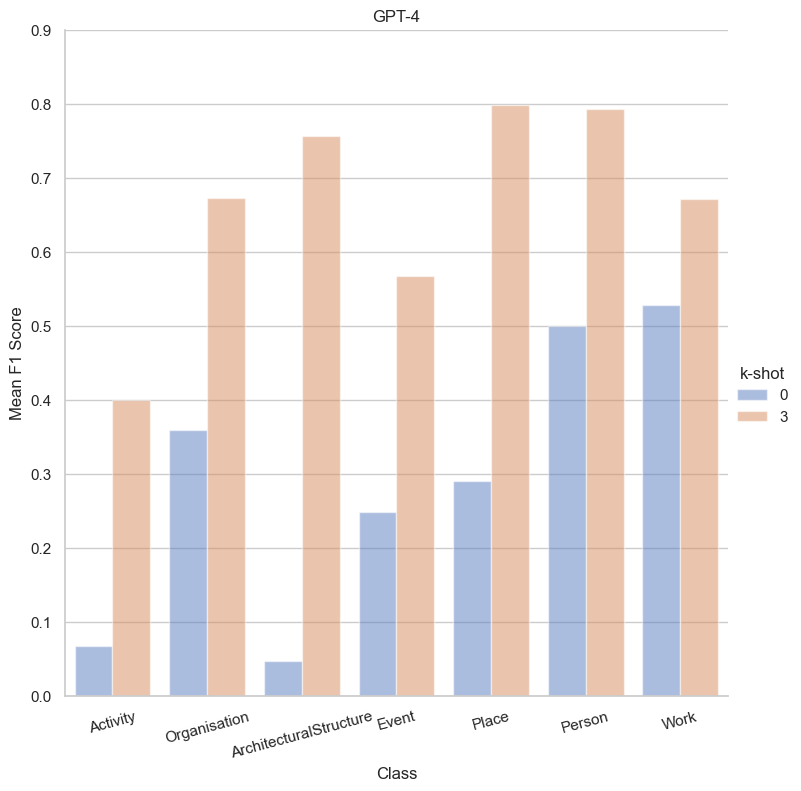}   
    \end{subfigure}
    \hspace{-1pt}
    \begin{subfigure}[b]{0.3\textwidth}
        \centering
        \includegraphics[width=\textwidth]{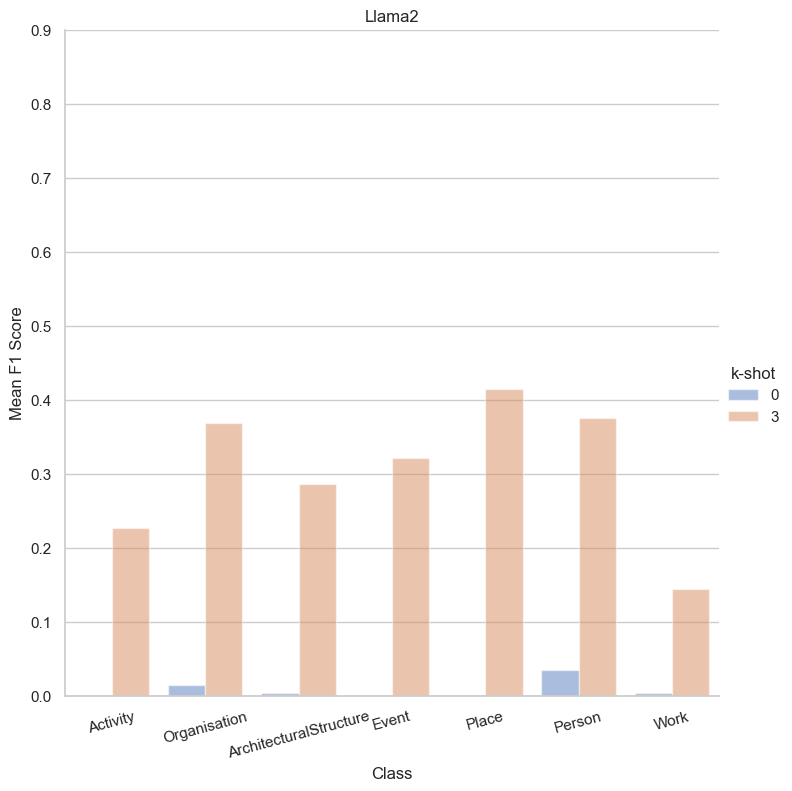}  
    \end{subfigure}
\vskip\baselineskip
\caption{Class-wise evaluation of the models 0 vs 3 shot examples sampled by similarity.}
\label{fig:class_wise}
\end{figure*}

\paragraph{\textbf{Class-wise evaluation}}
We show the class-wise results for the zero and three-shot experiments in Figure \ref{fig:class_wise}.
For all the models, the most challenging entities to annotate are those of type \textit{Activity}, while \textit{Person} is the type where all the models have the highest F1-score when not shown any examples. This difficulty could be attributed to the quantity of entities that are labeled with the Activity type, or the resemblance between entities of the Activity and Event type (as discussed in the Limitations section). The Llama2 model exhibited very low performance when not shown any examples, with the highest F1-score for the class \textit{Person} of just $0.035$. It is interesting to note, that in the case of 0-shot, the GPT-instruct is either on par or even better than the more powerful GPT-4 model. 
In Figure \ref{fig:class_wise}, we also see the improvement across all classes in the case of the three-shot experiment. We observe the most notable improvement in the class \textit{Architectural Structure}, where across all of the GPT models, has a significant increase in F1-score (from $\sim 0.04$ to $\sim 0.58$ for the GPT-3.5-turbo and GPT-4 models).

\section{Qualitative analysis}
In this section, we delve into a detailed qualitative analysis of the errors made by the LLMs, specifically focusing on errors in cell and span position and type prediction.

\paragraph{\textbf{Errors in Cell and Span Position}}

To assess the errors in cell position, we count the instances where entities have the correct span and label but are located in different cells compared to the ground truth. Our analysis reveals that less than 20\% of the erroneous predictions across the GPT models are from inaccuracies in cell positioning. For the LLama2 model, this figure rises to 36\% of incorrect predictions. In contrast, errors in predicting span positions are considerably less. Specifically, in less than 1\% of the incorrect predictions the model correctly identifies the cell position and label but misjudges either the beginning or end of the span.

\paragraph{\textbf{Errors in Type Prediction}}
We observe that another common mistake of the LLMs is to assign a too specific type for an entity. For example, the GPT-instruct model recognizes the entity ``Football'' as \textit{Sports}, instead of as \textit{Activity}. Even though this assignment is not wrong, the entity type \textit{Sports} is not included in the types given in the instruction part.
Table \ref{tab:predicted_types} shows an example of predicted entity types by the GPT-4 model and the GPT-instruct model in the case of zero-shot evaluation. The GPT-instruct model hallucinates more often and assigns more specific entity types then the types given in the instruction. It also invents entity types which are not specific types such as \textit{Edition, Founded} and \textit{Schools included}. The GPT-4 model mostly uses only the given types for annotation and invented only $2$ more specific type, \textit{Genre} and \textit{Job}. Similarly, the Llama2 model often assigns very specific types. In the zero-shot setting, it had predicted $123$ different entity types. However, with the 3-shot examples, sampled by similarity, this number is reduced to $54$.

\begin{table}
\centering
\caption{Predicted entity types by GPT-4 and GPT-instruct. The entities in red are hallucinated by the models.}
\label{tab:predicted_types}
\begin{tabular}{l l} 
\hline
\textbf{Model} & \textbf{Predicted Types} \\
\hline
\multirow{1}{*}{GPT-4} & \small Activity, Architectural Structure, Event \\ 
                       & \small\textcolor{red!80}{Genre}, \textcolor{red!80}{Job}, MISC, Organization, \\
                       & \small Person, Place, Work \\
                     
\hline
\multirow{1}{*}{GPT-instruct} & \small
Activity, \textcolor{red!80}{Administration  borough}, \textcolor{red!80}{Aircraft}, \\ 
                          & \small Arch. Struct.,  \textcolor{red!80}{Album}, \textcolor{red!80}{Band}, \textcolor{red!80}{Capacity}, \\
                          & \small \textcolor{red!80}{Centre of administration}, \textcolor{red!80}{Conference name}, \\
                          & \small \textcolor{red!80}{Country}, \textcolor{red!80}{Date}, \textcolor{red!80}{Dates}, \textcolor{red!80}{Edition}, \textcolor{red!80}{Episode},  \\
                          & \small Event, \textcolor{red!80}{Genre}, \textcolor{red!80}{Home city}, Job, MISC,\\
                          & \small Organisation,   \textcolor{red!80}{Other towns}, Person, Place,\\
                          &\small \textcolor{red!80}{Race team}, \textcolor{red!80}{Result}, \textcolor{red!80}{Schools included}, \textcolor{red!80}{Score},\\
                          & \small \textcolor{red!80}{Sports},  \textcolor{red!80}{Stadium}, \textcolor{red!80}{Team} \textcolor{red!80}{Time}, \textcolor{red!80}{Tribe}, Work, Year\\    
\hline
\end{tabular}
\end{table}

\section{Ablation study}
Building upon the findings from our qualitative analysis, we conducted an ablation study to assess the impact of reducing label specificity on the performance of our models and further understanding of model misclassifications.

\paragraph{\textbf{Label Specificity}} \label{appendix_ablation}
We observe that many of the errors are due to the possibility that one entity is of type Event or Activity. Therefore, we evaluate how much the models will improve if we keep only the 4 most distinct labels. We reduce the set of labels to : $\{Organisation, Place, Person, Work\}$. Table \ref{tab:filtered_res} shows the summarized results of the experiments with reduced label set. We observe that except for the Llama2 model, the overall performance of the rest of the models did not improve. Nevertheless, a comparison on a class level between the results on the reduced label dataset with and without any examples, demonstrates the increased ability of the model to correctly identify instances of the given classes. Even though the model still misclassifies instances of class \textit{Organization} as class \textit{Place}, the overall number of identified instances per class is doubled.

\begin{table}
    \centering
    \begin{tabular}{c c c c c c}
    \toprule
       & \multicolumn{2}{c}{\textbf{Random}} & \multicolumn{2}{c}{\textbf{Similarity}} & \\
        \midrule
         & 1-shot & 3-shot  & 1-shot & 3-shot   \\
         \midrule
       GPT-instruct & 0.51 & 0.52 & 0.68 & 0.71   \\
       GPT-3.5-turbo & 0.42 & 0.47 & 0.68 & \textbf{0.73}  \\
       GPT-4 & 0.43 & 0.47 & 0.69 & 0.71  \\
       Llama2-7b & 0.05 & 0.09 & 0.28 & 0.38  \\
       \bottomrule
    \end{tabular}
    \caption{F1-score with k-shot examples sampled at random and by similarity with reduced set of labels.}
    \label{tab:filtered_res}
\end{table}

\paragraph{\textbf{Type Prediction Analysis}}

\begin{figure}
    \centering
    \includegraphics[width=0.8\linewidth]{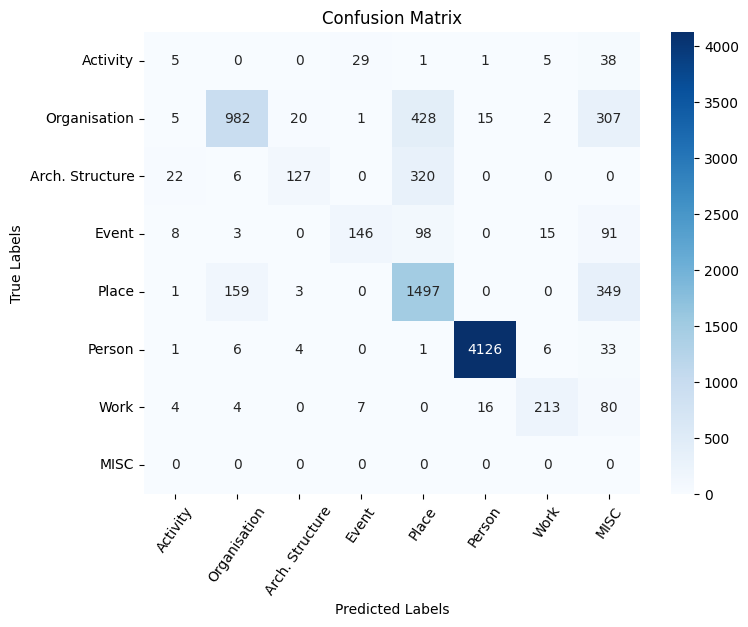}
    \caption{Confusion Matrix - GPT-instruct zero-shot}
    \label{fig:conf matrix}
\end{figure}

To better understand the misclassifications of the models, we also calculate the confusion matrix for the predictions. All the predicted types which are not part of the instruction, are represented as type \textit{MISC}. We show the confusion matrix of the GPT-instruct model with 0-shot examples in Figure \ref{fig:conf matrix}. The last row of the matrix is with zeros because we do not include any entities without labels in the ground truth. The last column of the matrix represents the number of entities per class that were classified as an unknown type (\textit{MISC}). We observe that the highest number of miss classification is for the entities of type \textit{Organization}; these instances are either misclassified as type \textit{Place} or as some other, unspecified type \textit{MISC}. 

\section{Limitations}\label{issues}
We now discuss the limitations of the proposed dataset and the data quality issues.

One issue identified with the qualitative analysis of the LLMs output is the datasets' lack of annotations. As illustrated in Figure \ref{fig:complex_table}, there are certain linked entities for which we fail to find their entity type, so these are labeled with 0 in the dataset. Remarkably, we noticed that the LLMs consistently assign correct labels to such entities. For instance, entities like ``Astor Piazzola'' and ``Antonio Carlos Jobim'' are accurately identified by all GPT models and labeled as \textit{Person}. 

Another issue is the existence of ambiguous entities, where multiple annotations are possible. This ambiguity often arises between entities categorized as \textit{Activity} and \textit{Event}. These are entities such as ``Winter Olympic Games 2010'' or ``2011 Cannes Festival'' which in different context can be an instance of both of the classes. On the one hand, since these involve various activities, such as sports competitions or film screenings, they can be viewed as an instance of the class \textit{Activity}. On the other hand, these are particular events with a particular duration. Thus, it can be difficult to define a class for such entities. 

Lastly, the annotations in the Wiki-TabNER dataset represent the most general class. We made this decision to avoid numerous small classes that could hinder the fine-tuning of transformer models which require larger train and test sets. Nevertheless, the LLMs showed to be capable of detecting the more fine-grained entity~types. 
Introducing more granular classification could better capture the nuances between overlapping entity labels. However,  deciding the level of granularity is challenging, as is ensuring that entities have corresponding labels in DBpedia (for instance, entity type Person, other type Athlete). 
Despite these limitations, the Wiki-TabNER dataset is sufficiently challenging for the new LLMs and a good starting point for evaluating the NER task in tables.

\section{Conclusion}
Recognizing the complexity of the tables in the real-world and using more challenging dataset for solving TI tasks is essential. This is because employing a more complex dataset not only reflects real-world scenarios more accurately, but also ensures that models are robust and effective in handling complex table structures. We demonstrated that using a more complex dataset requires the accurate recognition of NER in tables before addressing the other tasks. Our evaluation showed that NER within tables is a challenging task even for the current state-of-the-art LLMs. A solution to the NER in tables is the first step towards achieving a complete and correct information extraction from tables. 

The proposed dataset is documented and publicly available for download\footnote{\url{https://github.com/table-interpretation/wiki_table_NER}}. Its more than 200 downloads to date demonstrate its relevance and utility for the research community. By facilitating further advancements in information extraction from tables, this dataset serves as a valuable resource for ongoing and future studies.

As future work, we aim to extend the Wiki-TabNER dataset with additional tables and to improve the NER annotations. We are already exploring on expanding the evaluation to multi-label classification. This would allow entities to be associated with multiple labels from multiple levels (both Athlete and Person as correct labels) and ensure more fair evaluation of the LLMs. The current state of the dataset includes all the necessary info for expanding to multi-label classification (by expanding the subclasses of the current classes).
Furthermore, we plan to extend the evaluation to the EL task on the proposed dataset.
We hope we can stimulate interest in future research on the topic of table NER and facilitate these endeavors with the proposed dataset.

\bibliographystyle{ACM-Reference-Format}
\balance
\bibliography{custom}

\end{document}